\documentclass{article}

\usepackage{PRIMEarxiv}

\usepackage[utf8]{inputenc} % allow utf-8 input
\usepackage[T1]{fontenc}    % use 8-bit T1 fonts
\usepackage{hyperref}       % hyperlinks
\usepackage{url}            % simple URL typesetting
\usepackage{booktabs}       % professional-quality tables
\usepackage{amsfonts}       % blackboard math symbols
\usepackage{nicefrac}       % compact symbols for 1/2, etc.
\usepackage{microtype}      % microtypography
\usepackage{lipsum}
\usepackage{fancyhdr}       % header
\usepackage{graphicx}       % graphics
\usepackage{algorithm}
\usepackage{algorithmic}
\usepackage{multirow}
\usepackage{float}
\usepackage{anyfontsize}
\usepackage{siunitx}
\usepackage{capt-of}
\usepackage[labelfont=bf,labelsep=period]{caption}
\graphicspath{{media/}}     % organize your images and other figures under media/ folder

\title{DiTTO-LLM: Framework for Discovering Topic-based Technology Opportunities via Large Language Model}

\author{
  Wonyoung Kim \\
  Department of Library and Information Science \\
  Chung-Ang University \\
  Seoul, Korea 06974\\
  \texttt{kartboy10@cau.ac.kr} \\
   \And
  Sujeong Seo \\
  Insurance Strategy Planning Team \\
  Postal Savings \& Insurance Development Institute \\
  Seoul, Korea 07219\\
  \texttt{seos@posid.or.kr} \\
   \And
  Juhyun Lee\thanks{Corresponding Author} \\
  Future Technology Analysis Center \\
  Korea Institute of Science and Technology Information \\
  Seoul, Korea 02456\\
  \texttt{leeju@kisti.re.kr} \\
}

\begin{document}
\maketitle

\begin{abstract}
Technology opportunities are critical information that serve as a foundation for advancements in technology, industry, and innovation. This paper proposes a framework based on the temporal relationships between technologies to identify emerging technology opportunities. The proposed framework begins by extracting text from a patent dataset, followed by mapping text-based topics to discover inter-technology relationships. Technology opportunities are then identified by tracking changes in these topics over time. To enhance efficiency, the framework leverages a large language model to extract topics and employs a prompt for a chat-based language model to support the discovery of technology opportunities. The framework was evaluated using an artificial intelligence patent dataset provided by the United States Patent and Trademark Office. The experimental results suggest that artificial intelligence technology is evolving into forms that facilitate everyday accessibility. This approach demonstrates the potential of the proposed framework to identify future technology opportunities.
\end{abstract}

% keywords can be removed

\section{Introduction}
\label{ch1}
Song \textit{et al}., (2017) argued that technology opportunities discovery (TOD) involves anticipating technologies that are either emerging or in need of development \cite{Song2017}. They further emphasized that such opportunities have the potential to drive new market creation. In other words, technology opportunities can identify emerging technologies and stimulate innovation across markets and industries. Consequently, numerous countries and companies have pursued efforts to identify technology opportunities across various sectors.

Yoon \& Park, (2005) emphasized the importance of predicting both the direction and pace of technological development to identify technology opportunities effectively \cite{Yoon2005}. Even earlier, Porter \& Detampel, (1995) argued that TOD can support a variety of objectives \cite{Porter1995}. For example, technology opportunities can help identify (i) relationships among various technologies and (ii) how these technologies evolve over time. Therefore, by forecasting the trajectory and speed of technological advancements, it is possible to gain insights into the complex interconnections between technologies and track their evolution.

\begin{table}[H]
\centering
\caption{List of related works for TOD}
\label{table1}
\begin{tabular}{clccccccc}
\toprule
\textbf{Approaches} & \textbf{Related works} & \textbf{AHP} & \textbf{Score} & \textbf{Trends} & \textbf{Cluster} & \textbf{GTM} & \textbf{Class} & \textbf{Text} \\
\midrule
\multirow{3}{*}{Expert-based} 
& Cho \& Lee, (2013) & $\surd$ & & & & & & \\
& S. Lee \textit{et al}., (2014) & $\surd$ & $\surd$ & & & & & \\
& Y. Lee \textit{et al}., (2014) & $\surd$ & & & & & $\surd$ & \\
\midrule
\multirow{5}{*}{Time series-based}
& Kajikawa \& Takeda, (2008) & & & $\surd$ & $\surd$ & & & \\
& Kajikawa \textit{et al}., (2008) & & & $\surd$ & $\surd$ & & & \\
& Ogawa \& Kajikawa, (2015) & & & $\surd$ & & & & $\surd$ \\
& Lee \& Lee, (2019) & & $\surd$ & $\surd$ & & & $\surd$ & \\
& Lee \textit{et al}., (2020) & & $\surd$ & $\surd$ & & & & $\surd$ \\
\midrule
\multirow{4}{*}{Mapping-based}
& Yoon \textit{et al}., (2012) & & & & & $\surd$ & $\surd$ & \\
& Han \textit{et al}., (2021) & & & & & $\surd$ & $\surd$ & \\
& Kim \textit{et al}., (2020) & & & & & $\surd$ & $\surd$ & \\
& Liu \textit{et al}., (2023) & & & & & $\surd$ & $\surd$ & \\
\midrule
\multirow{4}{*}{Text-based}
& Yoon \& Magee, (2018) & $\surd$ & & & & $\surd$ & & $\surd$ \\
& Ren \& Zhao, (2021) & & & $\surd$ & & & & $\surd$ \\
& Wang \textit{et al}., (2023) & $\surd$ & & & & & & $\surd$ \\
& Zhao \textit{et al}., (2024) & & & & & $\surd$ & $\surd$ & $\surd$ \\
\bottomrule
\end{tabular}
\end{table}

More recently, Noh \textit{et al}., (2016) classified technology opportunities into four categories \cite{Noh2016}. The first type is technological vacancy, referring to underexplored technologies with high potential to create new markets as competition for development intensifies. The second type, convergent technology, encompasses technologies that are driving future advancements. The third type, emerging technology, represents opportunities to identify technologies poised for rapid growth. The final type, customer-based technology, is identified from a market perspective and focuses on technologies directly aligned with consumer needs.

In summary, TOD that drive innovation across technology, markets, and industries is of critical importance. However, as technology opportunities are diverse—ranging from convergent to emerging technologies—the discovery process is inherently challenging. Therefore, effectively discovering these opportunities requires a comprehensive understanding of the relationships among technologies and how they evolve over time.
Previous studies have explored various approaches to TOD. One such approach is expert-based, which relies on insights gathered from experts \cite{Cho2013,S_Lee2014,Y_Lee2014}. These approaches often utilize methods such as the Analytic Hierarchy Process (AHP) or Delphi surveys to systematically collect expert opinions on potential technology opportunities.

The second approach is time series-based, which aligns with the conventional understanding that TOD should consider temporal changes in technology \cite{Kajikawa_et_al_2008,Kajikawa_Takeda_2008,Lee2020,Lee2019,Ogawa2015}. From this perspective, time series-based approaches identify technology opportunities by tracking how technology clusters or keywords evolve over time.
Another approach is mapping-based, which involves projecting various technological attributes onto a two-dimensional space \cite{Han2021,Kim2020,Liu2023,Yoon2012}. This approach enables TOD by leveraging multidimensional information, allowing for a more comprehensive analysis of technological trends and relationships.

Recently, text-based approaches have gained prominence in TOD research, driven by improved access to large-scale text datasets, advancements in applicable algorithms, and enhanced performance of analytical models \cite{Ren2021,Wang2023,Yoon2018,Zhao2024}. Most studies in this area utilize datasets rich in technological information, such as academic publications and patents.
Previous studies on TOD can be categorized into expert-based, time series-based, mapping-based, and text-based approaches. This paper proposes a framework that combines time series, mapping, and text-based approaches to identify technology opportunities. In this framework, text-based topics are extracted from patent data to identify emerging technology opportunities. Technologies that have shown quantitative growth are identified through topic mapping, and topic trends are analyzed to track technologies that are increasing over time.

The contributions of this paper are as follows. First, topic mapping enables the identification of relationships among technologies, which aids in detecting technological vacancies and convergent technologies. Second, topic trends allow for tracking technological changes over time, facilitating the discovery of rapidly growing, emerging technologies. Finally, the proposed framework leverages the mapping and trend analysis of text-based topics to identify technology opportunities, specifically those exhibiting quantitative growth and temporal increase.

Chapter \ref{ch2} presents the fundamental concepts of large language model (LLM)-based topic modeling and reviews the TOD-related literature. Chapter \ref{ch3} introduces \texttt{DiTTO-LLM}, which consists of three main modules, and explains how it extends the literature discussed in Chapter \ref{ch2}. Chapter \ref{ch4} reports two experiments conducted to validate the proposed framework and its outcomes. The results of these experiments are discussed in Chapter \ref{ch5}. Finally, the limitations of our framework and directions for future research are elucidated in Chapter \ref{ch6}.

\section{Background} % Chapter 2
\label{ch2}
This paper identifies technology opportunities using topics generated from a LLM applied to an artificial intelligence (AI) patent dataset provided by the United States Patent and Trademark Office (USPTO). We begin by introducing the USPTO dataset and describing the LLM-based topic model. Finally, we review related work on TOD.

\subsection{USPTO AIPD} % 2.1
The USPTO is the federal agency responsible for granting U.S. patents and registering trademarks. The USPTO has released the Artificial Intelligence Patent Dataset (AIPD), which tracks the application of AI technology in U.S. patents issued between 1976 and 2020 \cite{Giczy2022,Toole2020}. The AIPD dataset, designed to support researchers and policymakers, enables the identification of specific AI technologies applied within patents.
The USPTO has categorized patents by dividing AI technology into eight distinct areas.

The eight AI technologies included in the AIPD are as follows:
\begin{itemize}
\item \textbf{Hardware}: Technologies related to physical computer components designed to enhance the computing power of AI algorithms by increasing processing efficiency and speed.
\item \textbf{Machine Learning (ML)}: Technologies involving computational models that learn from data.
\item \textbf{Natural Language Processing (NLP)}: Technologies for understanding and utilizing data encoded in written language.
\item \textbf{Speech}: Technologies focused on understanding sequences of words based on voice signals.
\item \textbf{Vision}: Technologies for extracting and interpreting information from visual inputs, including images and videos.
\item \textbf{Planning/Control}: Techniques for identifying and implementing strategies to achieve specific goals.
\item \textbf{Knowledge Processing}: Technologies for representing and deriving facts about the world for use in automated systems.
\item \textbf{Evolutionary Computation}: A field of optimization algorithms that simulate the evolutionary processes of living organisms.
\end{itemize}

The USPTO's AIPD enables identification of specific AI technologies applied within a patent. To develop the AIPD, the USPTO trained a prediction model to classify patents as AI-related based on two datasets: a Seeds set, representing positive examples of AI patents, and an Anti-Seeds set, representing negative examples.

The initial Seeds set was defined by linking the Examiners Automated Search Tool (EAST) to the USPTO database and further expanded by including patents linked through backward and forward citations, as well as patent families—patents with similar invention concepts.

Additional expansion of the Seeds set included patent family data, backward and forward citations, and cooperative patent classification (CPC) codes, culminating in a final Seeds set that also included citation families associated with the expanded patents.

The Anti-Seeds set, comprising patents unrelated to AI, was created by random sampling from patents outside the Seeds set due to its larger size, which reflects the broad range of non-AI technologies. Using a combined dataset of the Seeds and Anti-Seeds sets, the USPTO trained a prediction model to determine whether a patent pertains to one of the eight AI technology categories.

This model uses a patent’s text, backward citations, and forward citations as input vectors and outputs a probability, ranging from 0 to 1 through a sigmoid activation function, indicating the likelihood of AI relevance. Patents with an output of 0.5 or higher were classified as AI-related. The model's performance was validated by comparing its classifications to those made by patent examiners, achieving an accuracy of approximately 92\%.

\subsection{LLM-based topic model} % 2.2
Mohr \& Bogdanov, (2013) introduced the topic model as a method for automatically extracting meaningful content from large text corpora \cite{Mohr2013}. They argued that topic models enable the analysis of large-scale social phenomena that were previously unobservable due to the increasing volume of texts produced across various societal, scientific, and cultural contexts. A topic model is a NLP technique used to summarize extensive text collections into distinct themes or topics.

Papadimitriou \textit{et al}., (1998) proposed latent semantic analysis (LSA) for topic extraction from text \cite{Papadimitriou1998}. LSA transforms large text corpora into a document-term matrix, applying singular value decomposition to yield a semantic matrix representing the document’s core topics. Subsequently, Blei \textit{et al}., (2003) introduced latent Dirichlet allocation (LDA), which models topics by considering the probability of topic occurrence \cite{Blei2003}. LDA assumes that topics generate words based on specific probability distributions, enabling the estimation of topic distributions within large-scale texts.

Recently, LLMs have been actively explored for topic extraction, as they excel at various NLP tasks, including document summarization. Grootendorst, (2022) proposed BERTopic, a topic model that utilizes Bidirectional Encoder Representations from Transformers (BERT) \cite{Grootendorst2022}.

BERTopic extracts topics through a five-step process:

\begin{itemize}
\item BERTopic first converts large-scale text into vectors using BERT, enabling the measurement of similarity between texts in vector form.
\item BERTopic then reduces the dimensionality of these text vectors to facilitate efficient processing and to enhance the identification of semantically similar texts.
\item In the third step, semantically similar texts are grouped together through clustering, with each cluster representing a topic.
\item Next, BERTopic extracts representative terms for each cluster by calculating term frequency-inverse document frequency (TF-IDF) scores. Topics are defined based on terms with high TF-IDF values within each cluster.
\item Finally, instead of simply ranking terms by TF-IDF, BERTopic uses maximum marginal relevance (MMR) to extract a diverse set of representative terms for each topic. MMR ensures that the selected terms capture a variety of aspects within each topic.
\end{itemize}

BERTopic offers flexibility in applying various BERT-based LLMs to different contexts with ease and efficiency. For instance, BERTopic can utilize a version of BERT fine-tuned specifically for patent contexts, allowing for targeted topic extraction from patent documents.

\subsection{Related works for TOD} % 2.3
Related work on TOD can be categorized into four main approaches. Expert-based approaches identify technology opportunities by gathering expert opinions, often through methods such as the AHP or Delphi surveys \cite{Cho2013}. These approaches have also introduced two-stage TOD processes that incorporate quantitative scoring \cite{S_Lee2014} or integrate subject-action-object (SAO) structures to mitigate potential biases arising from experts' subjective opinions \cite{Y_Lee2014}.

Time series-based approaches identify technology opportunities by examining temporal changes in technological attributes. For example, Kajikawa \& Takeda, (2008) \cite{Kajikawa_Takeda_2008} and Kajikawa \textit{et al}., (2008) \cite{Kajikawa_et_al_2008} tracked shifts in technology clusters over time to reveal emerging opportunities. More recent time series-based approaches have focused on monitoring changes in keywords \cite{Ogawa2015} or scores \cite{Lee2020,Lee2019} for technical attributes over time.

Mapping-based approaches commonly employ generative topographic mapping (GTM) for TOD. For instance, Liu \textit{et al}., (2023) constructed a hierarchical semantic network using SAO structures and applied GTM to identify relationships between technologies \cite{Liu2023}. These approaches are particularly effective in discovering inter-technology relationships \cite{Han2021,Kim2020,Yoon2012}.

Text-based approaches conduct TOD using technological documents such as academic publications and patents. For example, B. Yoon \& Magee, (2018) \cite{Yoon2018} identified technology opportunities through GTM based on keywords, while Ren \& Zhao, (2021) \cite{Ren2021} detected technological changes via time series regression on keywords. Recently, TOD using LLM has also been attempted. Wang \textit{et al}., (2023) \cite{Wang2023} proposed a TOD method that combines BERT-based topic extraction with the Theory of Inventive Problem Solving (TRIZ), while Zhao \textit{et al}., (2024) \cite{Zhao2024} introduced a technique for extracting topics based on SAO structures. Text-based approaches offer an efficient means to understand inter-technology relationships through keywords or topics.

Table \ref{table1} summarizes related works on TOD, highlighting the use of expert-based, time series-based, mapping-based, and text-based approaches. However, existing research has limitations, as it often fails to account for both the relationships between technologies and their evolution over time. To address these gaps, this paper proposes a framework that integrates time series, mapping, and text-based approaches to enhance the discovery of technology opportunities.

\section{Proposed Methodology} % Chapter 3
\begin{figure}[H]
\centering
\includegraphics[scale=0.6]{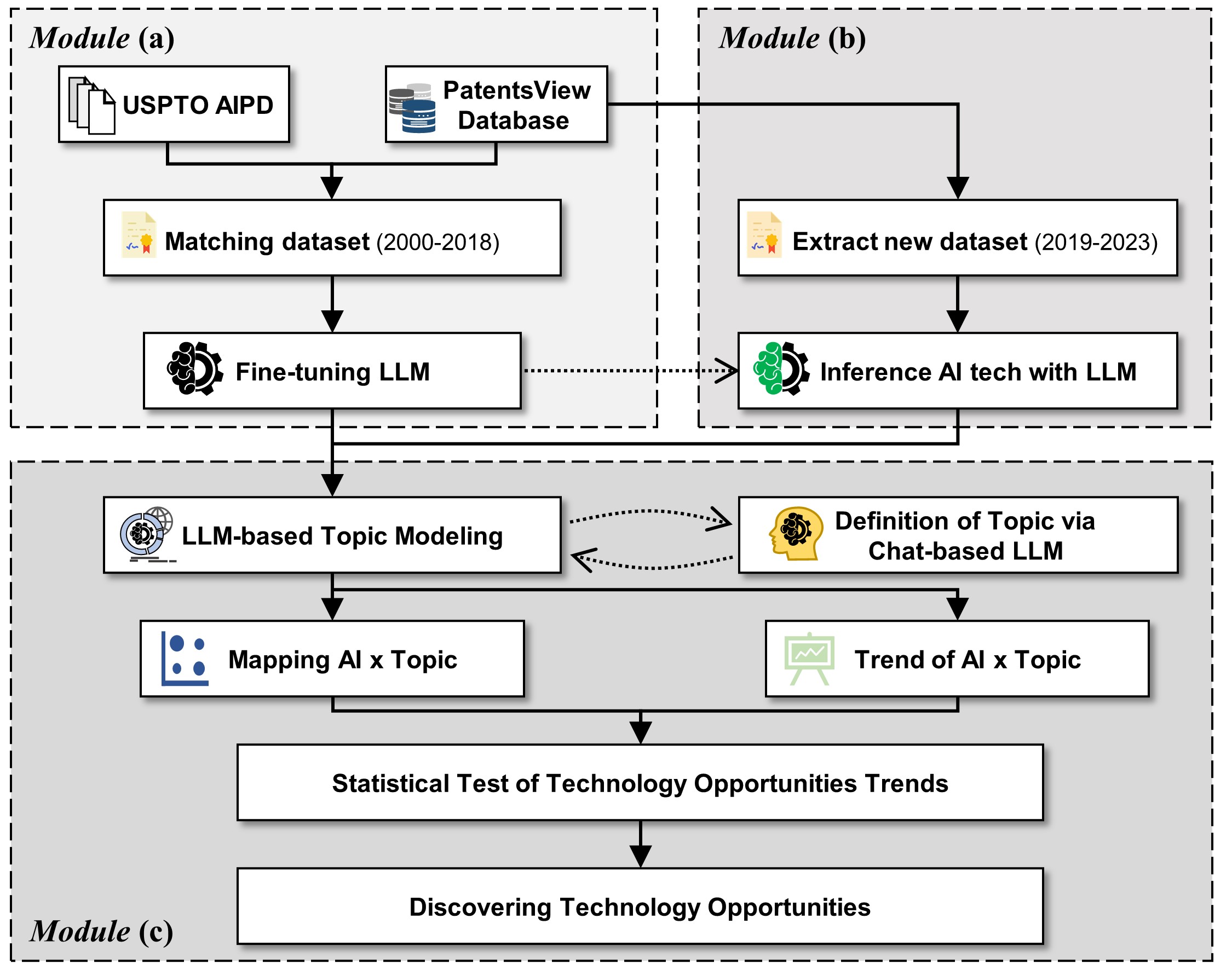}
\caption{Flowchart of the \texttt{DiTTO-LLM}.}
\label{fig1}
\end{figure}

\label{ch3}
This paper introduces a topic-based methodology for TOD in the field of AI using LLMs. We refer to this framework as \texttt{DiTTO-LLM} (Discovering Topic-based Technology Opportunities via LLM).
\texttt{DiTTO-LLM} consists of three modules, as illustrated in Figure \ref{fig1}. Module (a) involves matching the AIPD with the PatentsView database, resulting in a dataset of patents registered from 2000 to 2018. This matched dataset is categorized into six labels, which are then used to fine-tune the LLM for further analysis.
Module (b) extracts new datasets registered after 2018. Then, the label of the new dataset is inferred using the fine-tuned LLM in module (a). This is because AIPD only includes patents registered until 2018.
Lastly, module (c) discovers technology opportunities in the AI field from the patent dataset. This module uses fine-tuning LLM-based topic modeling. In this process, chat-based LLM is used to define topics in the AI field. Then technology opportunities are discovered through quantitative growth and time-series increase in topics and AI labels.

\subsection{Matching AI dataset and fine-tuning LLM} % 3.1
In module (a) of \texttt{DiTTO-LLM}, the AIPD is matched with the PatentsView database to facilitate LLM fine-tuning. AIPD provides a list of AI patents up to 2018, from which we extract the patent texts using the PatentsView database. These extracted texts are used to fine-tune the LLM.

\begin{center}
\begin{minipage}{0.6\textwidth}
\begin{algorithm}[H] % or 'htbp' (here, top, bottom, page)
\caption{Extract text from matching dataset}
\label{alg1} % Table 2 -> Algorithm 1
\begin{algorithmic}[1]
\STATE \textbf{Input:} AIPD: $D_a$, PatentsView Database: $D_p$, Query: $Q$
\STATE \textbf{Output:} Matching Dataset: $X_{ap}$
\STATE Set empty list $d_{ap}$
\FOR{$d_a$ in $D_a$}
    \IF{$d_a$ in $D_p$}
        \STATE $d_{ap} \leftarrow d_a$
    \ENDIF
\ENDFOR
\STATE $X_{ap} \leftarrow Q(d_{ap})$
\end{algorithmic}
\end{algorithm}
\end{minipage}
\end{center}

Algorithm \ref{alg1} provides the pseudo code for extracting text from the matching dataset. We denote the AIPD and PatentsView database as $D_a$ and $D_p$, respectively. First, module (a) identifies $d_{ap}$, which represents the intersection of $D_a$ and $D_p$ (lines 3--6); in other words, $d_{ap}$ is a list of AI patents registered up to 2018.

Module (a) then applies query $Q$ to extract the matching dataset $X_{ap}$ for analysis (line 9). This query retrieves variables from $d_{ap}$ in the PatentsView database. Table A1 in the Appendix provides a sample query that extracts the matching dataset $X_{ap}$, including fields such as title and abstract.

\begin{center}
\begin{minipage}{0.8\textwidth}
\begin{algorithm}[H]
\caption{Define labels for matching dataset}
\label{alg2} % Table 3 -> Algorithm 2
\begin{algorithmic}[1]
\STATE \textbf{Input:} Matching Dataset: $X_{ap}$, Threshold of probability: $\gamma$, $i$-th AI label: $AI^{(i)}$
\STATE \textbf{Output:} List of AI labels: $Y_{ap}$
\FOR{$x_{ap}$ in $X_{ap}$}
    \STATE $P_{ap} \leftarrow x_{ap}$
    \STATE $P^{(i)} \leftarrow \arg\max(P_{ap})$
    \IF{$P^{(i)} \geq \gamma$}
        \STATE $Y_{ap} \leftarrow AI^{(i)}$
    \ELSE
        \STATE $Y_{ap} \leftarrow \textit{not}\_AI$
    \ENDIF
\ENDFOR
\end{algorithmic}
\end{algorithm}
\end{minipage}
\end{center}

Module (a) extracts the matching dataset $X_{ap}$ from the AIPD and PatentsView database. To fine-tune the LLM for classifying patents according to the technology categories defined by AIPD—Hardware, ML, NLP, Speech, and Vision—we must first assign labels to the entries in $X_{ap}$.

Algorithm \ref{alg2} provides the pseudo code for assigning labels to $X_{ap}$. Let $x_{ap}$ represent a patent within $X_{ap}$. Here, $P_{ap}$ is a five-dimensional list containing the probabilities that $x_{ap}$ belongs to each of the categories: Hardware, ML, NLP, Speech, and Vision. The label of $x_{ap}$ is determined by the argument of the maximum function (argmax) (lines 3--5). If the argmax value exceeds a probability threshold $\gamma$ , the label of $x_{ap}$ is assigned to the $i$-th AI category in $P_{ap}$; otherwise, it is labeled as \textit{not}\_AI (lines 6--11).

The contributions of fine-tuning are as follows:
\begin{itemize}
\item Fine-tuning enables the LLM to develop a deeper understanding of AI technology, facilitating the efficient use of an optimized LLM to identify technology opportunities in AI.
\item The fine-tuned LLM can infer labels for new datasets in module (b), capturing not only the five AI categories but also distinguishing non-AI technologies (\textit{not}\_AI).
\item In module (c), the fine-tuned LLM supports topic modeling, aiding in the extraction of topics that are particularly relevant to AI.
\end{itemize}

The purpose of module (a) is to compile a dataset and fine-tune the LLM for validating the proposed method. Since the matching dataset from module (a) consists of patents registered up to 2018, the resulting analysis reflects past technology opportunities. This allows us, from a vantage point beyond 2018, to assess the appropriateness and accuracy of these results.

\subsection{Extraction of a new dataset and inference label via LLM} % 3.2
Module (b) of \texttt{DiTTO-LLM} extracts patents from 2019 onward from the PatentsView database and infers AI labels for them. This module specifically retrieves datasets of patents registered between 2019 and 2023 to discover emerging technology opportunities in the AI field.

We use the new dataset obtained in module (b) to identify technology opportunities emerging after 2023 through the proposed methodology. To achieve this, we need to infer AI labels for the new dataset, as the AIPD only includes patents registered prior to 2018.

We use the fine-tuned LLM from module (a) to infer AI labels for the new dataset. The fine-tuned LLM outputs the probability that a patent belongs to one of the categories: Hardware, ML, NLP, Speech, Vision, or \textit{not}\_AI. Let $\tilde{x}_{ap}$ represent a patent in the new dataset. The output from the fine-tuned LLM is an $L$-dimensional vector, where $L$ is the number of AI labels. The $l$-th element $\tilde{P}^{(l)}_{ap}$ of $\tilde{x}_{ap}$, which corresponds to the $l$-th output $\tilde{z}^{(l)}_{ap}$ of the fine-tuned LLM, is calculated as shown in equation (\ref{eq1}).

\begin{equation}
  \tilde{P}^{(l)}_{ap}
  = \frac{e^{\tilde{z}^{(l)}_{ap}}}{\sum_{k=1}^{L} e^{\tilde{z}^{(k)}_{ap}}}
  \label{eq1}
\end{equation}

Next, the label of $\tilde{x}_{ap}$ is determined according to the pseudo code (lines 5--10) of Algorithm \ref{alg2}.

\subsection{Discovering technology opportunities via LLM-based topic model} % 3.3
Module (c) of \texttt{DiTTO-LLM} identifies topic-based technology opportunities. First, it constructs an LLM-based topic model, which discovers specific technical topics from the titles and abstracts of patents. The topics generated by this model represent technologies that integrate aspects of patents related to Hardware, ML, NLP, Speech, and Vision, highlighting areas of complex technological fusion.

\begin{table}[H]
\centering
\caption{Prompt that defines keywords-based technology opportunities}
\label{table2} % Table 4 -> Table 2
\begin{tabular}{p{0.03\textwidth} p{0.93\textwidth}}
\toprule
\textbf{Prompt} & \\ 
\midrule
1: & You are an expert in science and technology information. You can predict promising technologies through papers and patents. \\[0.5em]

2: & I'll give you a list of keywords. You look at it and name it emerging technology. Let me show you an example. \\[0.5em]

3: & \textbf{Keyword list:} wind, energy, hydrogen, electricity, pv, fuel, power, cost, production, turkey, demand, system, generation, storage, sources, ltd, battery, turbine, speed, diesel \\
   & \textbf{Related technology:} renewable energy storage and conversion technology utilizing hydrogen energy \\[0.5em]

4: & \textbf{Keyword list:} heat, air, temperature, adsorption, cooling, absorption, transfer, water, mass, cop, performance, desiccant, refrigeration, degrees, flow, cycle, system, chiller, exchanger, ltd \\
   & \textbf{Related technology:} next-generation eco-friendly heating and cooling system core material technology \\[0.5em]

5: & \textbf{Keyword list:} co2, reaction, catalyst, co, catalysts, reduction, carbonate, carbon, activity, cu, dioxide, tio2, synthesis, selectivity, formation, conversion, complexes, acid, dmc, carbonates \\
   & \textbf{Related technology:} carbon dioxide capture and storage technology \\[0.5em]

6: & \textbf{Keyword list:} vehicle, tire, road, wheel, control, tyre, steering, driver, braking, vehicles, friction, controller, model, yaw, stability, slip, brake, angle, force, dynamics \\
   & \textbf{Related technology:} advanced autonomous vehicle technology \\[0.5em]

7: & \textbf{Keyword list:} fruit, color, vision, imaging, images, image, quality, wavelengths, fruits, nm, features, processing, shape, meat, machine, samples, food, colour, detection, crop \\
   & \textbf{Related technology:} ai-based machine vision technology \\[0.5em]

8: & Do you understand how to do it? \\
\bottomrule
\end{tabular}
\end{table}

Subsequently, module (c) of \texttt{DiTTO-LLM} employs a chat-based LLM to assign names to the identified topics. Recently, chat-based LLMs such as GPT4 \cite{OpenAI2023} and GPT-oss \cite{Agarwal2025} have gained attention for their capability to address human inquiries by leveraging extensive pre-trained data. These models also excel at tasks like summarizing large documents, which can be challenging for humans due to the sheer volume of information. Accordingly, \texttt{DiTTO-LLM} utilizes a chat-based LLM to effectively define names for each topic.

Table \ref{table2} provides a prompt designed to guide the chat-based LLM in defining topics. The prompt instructs the chat-based LLM to examine keywords and identify technology opportunities. First, it assigns specific tasks and roles to the chat-based LLM (lines 1--2). Next, the prompt provides the LLM with information on technology opportunities \cite{Lee2021} along with keywords extracted from prior studies (lines 3--7). Finally, it includes a step to confirm the LLM's comprehension of the task (line 8).

\begin{center}
\begin{minipage}{0.7\textwidth}
\begin{table}[H]
\centering
\caption{Prompt to define technology opportunities in \texttt{{DiTTO-LLM}} results}
\label{table3} % Table 5 -> Table 3
\begin{tabular}{p{0.95\textwidth}}
\toprule
\textbf{Prompt} \\
\midrule
1: I'll show you the keyword list: \texttt{\{KEYWORD\_LIST\}}. \\[0.5em]
2: Now, give me a list of 5 technology opportunities. \\
\bottomrule
\end{tabular}
\end{table}
\end{minipage}
\end{center}

Table \ref{table3} presents a prompt for defining technology opportunities based on the output of \texttt{DiTTO-LLM}. In this prompt, \texttt{KEYWORD\_LIST} represents the keywords associated with each topic, as derived from module (c) of \texttt{DiTTO-LLM} (line 1). The chat-based LLM returns five lists, enabling \texttt{DiTTO-LLM} to assign suitable names to each identified technology opportunity (line 2).

Finally, \texttt{DiTTO-LLM} identifies technology opportunities based on two key characteristics. First, technology opportunities are those technologies that have shown continuous recent growth. Second, they exhibit quantitative expansion over time. To quantify these characteristics, we analyze time series trends and measure quantitative growth according to topic and AI label.

Let the topic of patent $x_{ap}$ be denoted as $t_{ap}$. The variation in $t_{ap}$ over the patent filing dates forms a time series dataset, allowing us to observe the trend of the topic over time. We analyze this dataset to confirm any increasing trend in the topic.

Let the AI label of patent $x_{ap}$ be denoted as $y_{ap}$. In module (c) of \texttt{DiTTO-LLM}, we compare $t_{ap}$ and $y_{ap}$ for each patent $x_{ap}$, a process we define as AI x Topic mapping. This mapping facilitates the identification of specific technologies (or topics) that are experiencing quantitative growth within each AI domain.

Finally, we examine the time series trend of the AI x Topic mapping, as a trend showing quantitative growth is more significant than a simple upward trend over time. In other words, \texttt{DiTTO-LLM} identifies AI x Topics that exhibit both quantitative growth and an upward time series trend as technology opportunities. Let $s^{(t)}_{ap}$ denote the set of technologies with topic $t_{ap}$ and AI label $y_{ap}$ at time $t$ (t=1,2,…,T). The linear growth trend $\beta_{ap}$, estimated using the averages $\bar{T}$ of $t$ and $\bar{s}_{ap}$ of $s^{(t)}_{ap}$, can be obtained as shown in equation (\ref{eq2}).

\begin{equation}
\frac{\sum_{t=1}^{T} (t - \bar{T}) \times \left( s^{(t)}_{ap} - \bar{s}_{ap} \right)}
  {\sum_{t=1}^{T} (t - \bar{T})^2}
\label{eq2}
\end{equation}

Equation (\ref{eq3}) presents the research hypothesis, which assumes that technology opportunities exhibit both quantitative growth over time $t$ and an increasing trend in the time series.

\begin{equation}
H_{0} : \beta_{ap} \leq 0 \;\; \textit{vs.} \;\; H_{1} : \beta_{ap} > 0
\label{eq3}
\end{equation}

Therefore, \texttt{DiTTO-LLM} defines $s^{(t)}_{ap}$, which rejects the null hypothesis at a specified significance level, as a technology opportunity. \texttt{DiTTO-LLM} is structured in three modules, (a) through (c). Modules (a) and (c) are used to fine-tune the LLM and validate the performance of \texttt{DiTTO-LLM}. Modules (b) and (c) are then applied to identify future technology opportunities by analyzing recently registered patents.

\renewcommand{\arraystretch}{1.2}
\begin{table}[h]
\centering
\begin{minipage}{0.7\textwidth}
\caption{AIPD used in experiments}
\label{table4} % Table 6 -> Table 4
\centering
\sisetup{table-format=5.0, table-number-alignment=center}
\begin{tabular}{c r S[table-format=2.2]}
\toprule
\textbf{Technology} & \textbf{Count} & \textbf{Rate (\%)} \\
\midrule
Hardware & 63,575  & 28.85 \\
ML       & 22,570  & 10.24 \\
NLP      & 7,630   & 3.46  \\
Speech   & 14,307  & 6.49  \\
Vision   & 89,735  & 40.72 \\
\textit{not}\_AI  & 22,570  & 10.24 \\
\bottomrule
\end{tabular}
\end{minipage}
\end{table}
\renewcommand{\arraystretch}{1.0} % return to default

\section{Experiments} % Chapter 4
\label{ch4}
The experimental results are presented in three subchapters. First, the experimental setup describes module (a) of \texttt{DiTTO-LLM}. The second subchapter identifies technology opportunities by applying the matching dataset, extracted from module (a), to module (c).

\begin{figure}[h]
\centering
\includegraphics[scale=0.5]{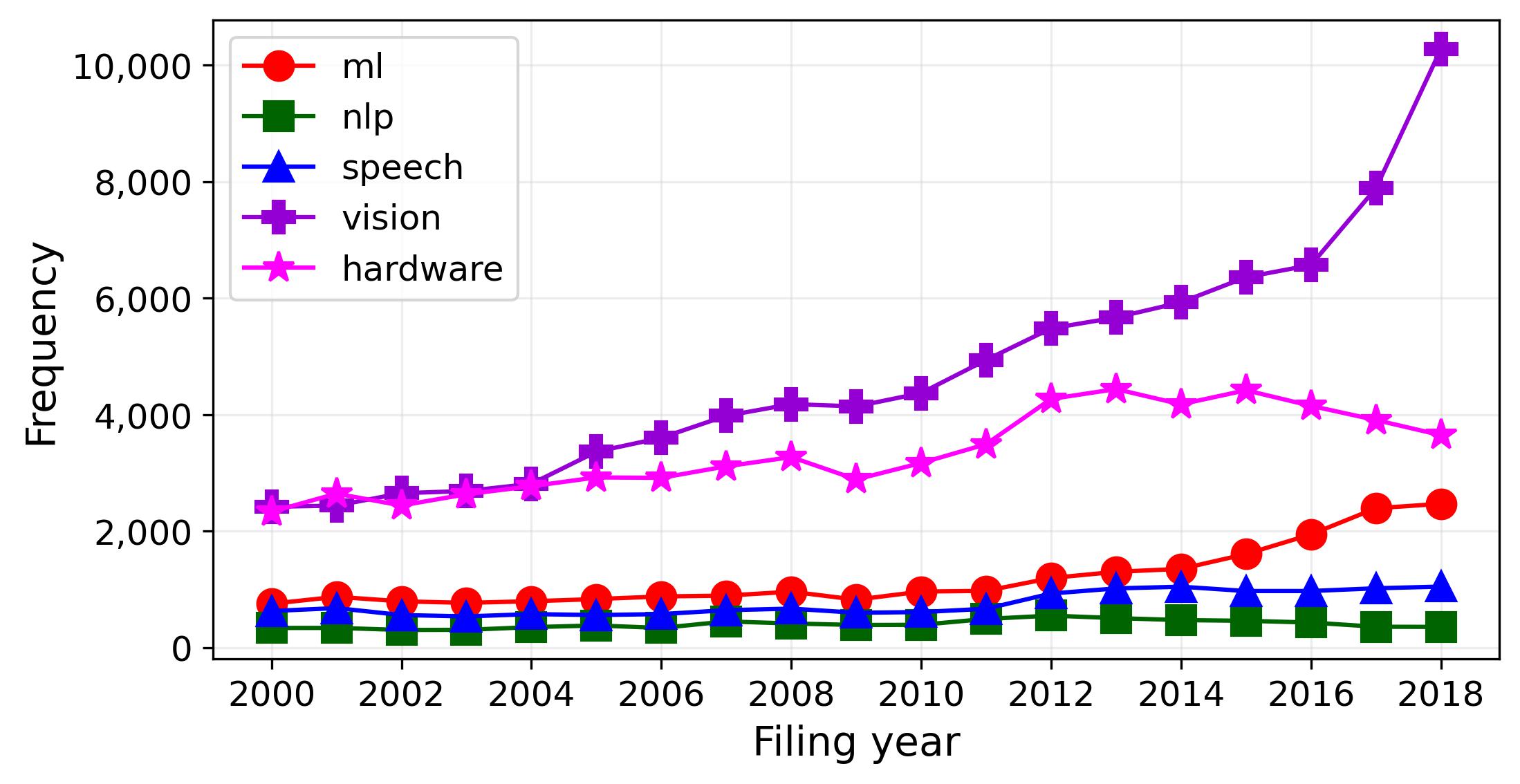}
\caption{Trends in the AIPD by filing year.}
\label{fig2}
\end{figure}

Since this dataset comprises patents from 2000 to 2018, this step allows us to examine the feasibility of \texttt{DiTTO-LLM}. The final subchapter presents technology opportunities discovered from the new dataset obtained in module (b). This experimental design enables a comprehensive evaluation of \texttt{DiTTO-LLM}, from validating its feasibility to discovering future technology opportunities.

\begin{figure}[H]
\centering
\includegraphics[scale=0.8]{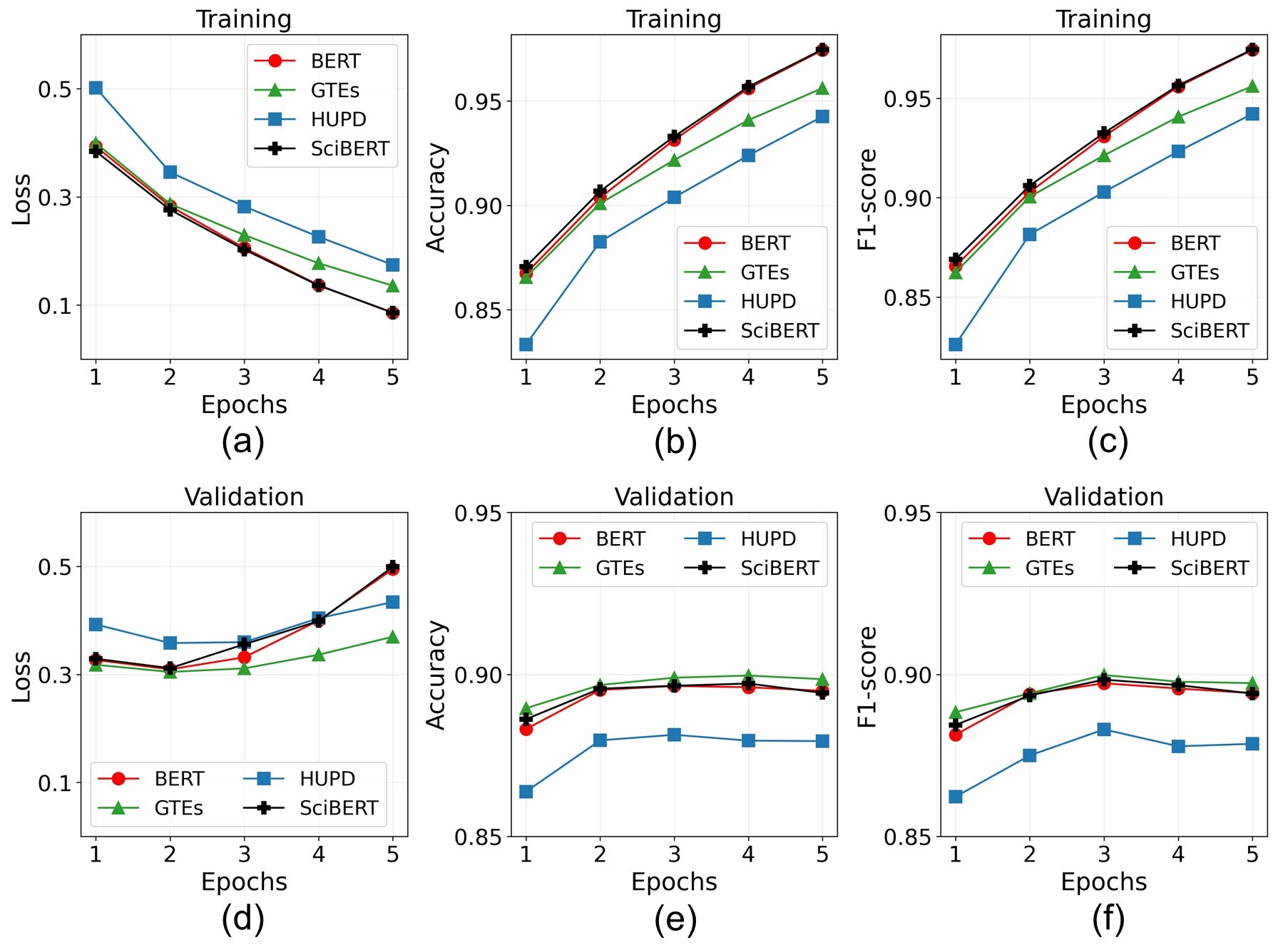}
\caption{Process of fine-tuning for each LLM.}
\label{fig3}
\end{figure}

\subsection{Module (a): Experimental setup} % 4.1
The data used in the experiment was extracted from the AIPD and PatentsView database. We selected patents related to Hardware, ML, NLP, Speech, and Vision from AIPD to identify AI technology opportunities. Additionally, to classify non-AI technologies, we included a sample of patents labeled as \textit{not}\_AI. Table \ref{table4} presents the 220,387 AIPD patents used in the experiment.

Figure \ref{fig2} illustrates the annual distribution of AIPD patents. The collected AIPD dataset includes patents registered between 2000 and 2018. Among AI technologies, Vision patents are the most prevalent and show a continuous upward trend. In contrast, Hardware patents—the second most common category—have declined since 2015. Additionally, NLP and Speech patents display a slight upward trend, though their counts remain lower than those of other AI technologies.

\renewcommand{\arraystretch}{1.2}
\begin{table}[H]
\centering
\begin{minipage}{0.8\textwidth}
\caption{List of topics extracted from matching dataset}
\label{table5} % Table 7 -> Table 5
\centering
\sisetup{table-number-alignment = center, table-format=3.2} % align float
\begin{tabular}{clS}
\toprule
\textbf{Topic} & \textbf{Description} & \textbf{Rate (\%)} \\
\midrule
T0 & AI-based Image Recognition System      & 38.02 \\
T1 & Edge Computing Technology              & 31.42 \\
T2 & Deep Learning Technology               & 13.88 \\
T3 & Multi-modal Data Processing            & 9.67  \\
T4 & Image-to-Text Converter                & 6.63  \\
T5 & Advanced Driver-Assistance Systems     & 0.22  \\
T6 & AI Hardware Accelerators               & 0.16  \\
\bottomrule
\end{tabular}
\end{minipage}
\end{table}
\renewcommand{\arraystretch}{1.0} % return to default

LLM fine-tuning was conducted using six patent labels: Hardware, ML, NLP, Speech, Vision, and \textit{not}\_AI. The dataset was split with a 9:1 ratio for training and validation, and fine-tuning was performed over five epochs, saving the model with the best performance. Four LLMs were employed in the comparison experiment. BERT \cite{Devlin2019}, the foundational model for LLMs, served as the baseline. GTE \cite{Li2023} is a model optimized for plain text embedding, and the small version was used in the experiment. HUPD \cite{Suzgun2024} is an LLM further trained on patent datasets, while SciBERT \cite{Beltagy2019} is an LLM pre-trained on scientific papers and patents.

Figure \ref{fig3} presents performance metrics (Loss, Accuracy, and F1-score) across epochs. The F1-score is the macro-average across the six labels. Figure \ref{fig3}(a)-(c) illustrate these metrics on the training dataset, while Figure \ref{fig3}(d)-(f) show them on the validation dataset. Comparing fine-tuning results on both datasets, GTEs achieved the best performance at three epochs. Therefore, fine-tuned GTEs are used in subsequent experiments.

\begin{figure}[H]
\centering
\includegraphics[scale=0.85]{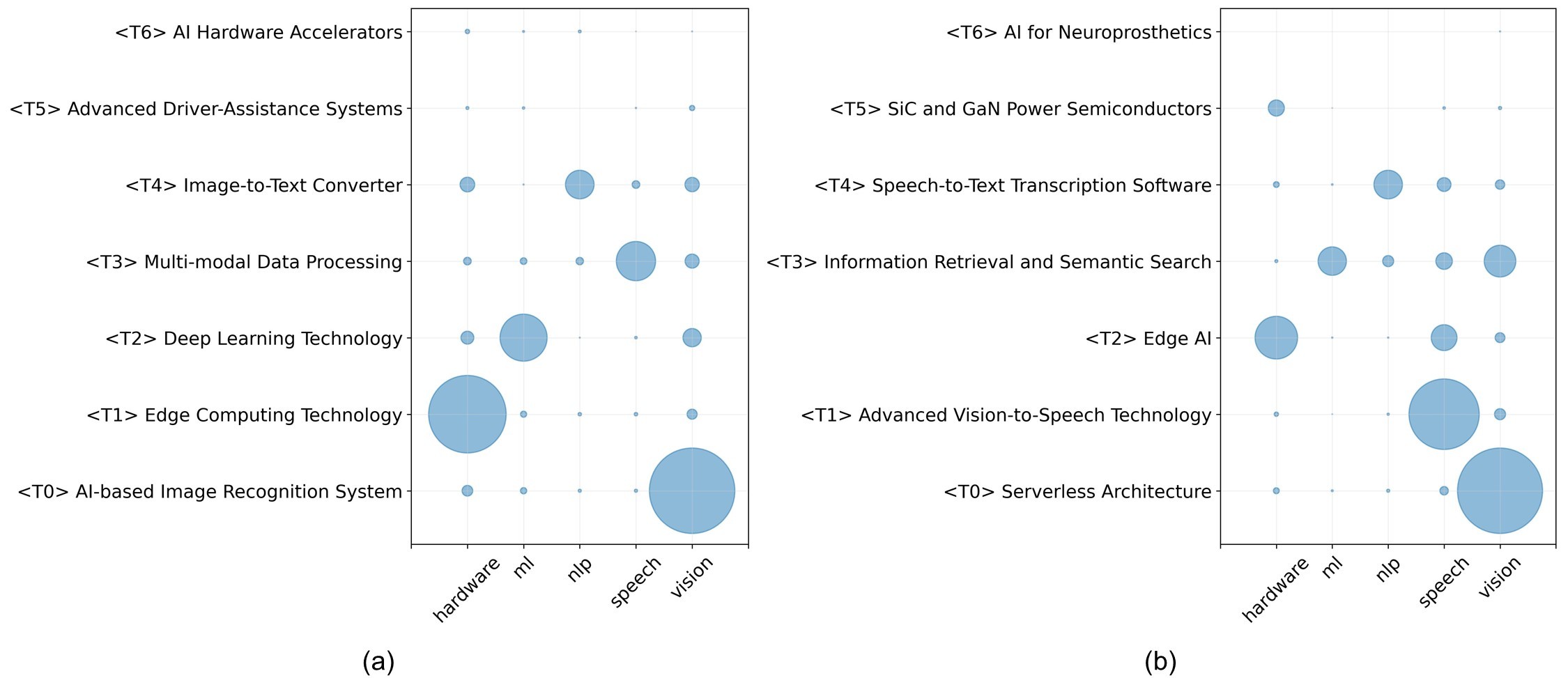}
\caption{Results of mapping AI x topic. (a) and (b) depict the results of matching dataset and new dataset, respectively.}
\label{fig4}
\end{figure}

\subsection{Module (a)+(c): TOD in 2000-2018} % 4.2
Now, topics were extracted using the matching dataset, and the LLM-based topic model was built with GTEs. Table \ref{table5} provides an overview of the identified topics.

Topic names were defined by applying the prompts from Table \ref{table2} and Table \ref{table3} in ChatGPT-4o. The topics identified from patents registered up to 2018 (matching dataset) can be categorized into data processing (T0, T2, T3), hardware configuration for commercialization (T1, T6), and business applications (T4, T5).

Figure \ref{fig4}(a) displays the mapping results of topics and AI labels within the matching dataset. As a result, Vision x T0 and Hardware x T1 may be saturated. In contrast, \{ML, Vision\} x T2 and Speech x T3 are of moderate size relative to other AI x topics, indicating potential for quantitative growth.

However, quantitative growth alone is insufficient to represent technology opportunities, necessitating an examination of the time series trends for AI x topics to identify true opportunities.

\begin{table}[H]
\centering
\begin{minipage}{0.95\textwidth}
\caption{Results of statistical tests for matching dataset}
\label{table6} % Table 8 -> Table 6
\renewcommand{\arraystretch}{1.2}
\begin{tabular}{clrrrrrrr}
\toprule
\textbf{Technology} & \textbf{Statistics} & \textbf{T0} & \textbf{T1} & \textbf{T2} & \textbf{T3} & \textbf{T4} & \textbf{T5} & \textbf{T6} \\
\midrule
\multirow{3}{*}{Hardware}
 & Coefficient & -2.800 & -99.800 & -4.800 & -2.100 & -3.000 & -0.100 & -1.100 \\
 & Std.Err     &  0.383 &  27.628 &  1.039 &  0.486 &  1.249 &  0.191 &  0.790 \\
 & T-value     & -7.311 &  -3.612 & -4.619 & -4.317 & -2.402 & -0.522 & -1.393 \\
\midrule
\multirow{3}{*}{ML}
 & Coefficient & \textbf{3.800} & -0.600 & \textbf{104.500} &  4.800 & -0.200 & -0.200 &  0.100 \\
 & Std.Err     & \textbf{1.442} &  0.462 & \textbf{22.370}  &  2.561 &  0.115 &  0.163 &  0.153 \\
 & T-value     & \textbf{2.635}$^{*}$ & -1.299 & \textbf{4.671}$^{\dagger}$ & 1.874 & -1.732 & -1.225 & 0.655 \\
\midrule
\multirow{3}{*}{NLP}
 & Coefficient &  0.400 & -1.100 &   --   & -1.600 & -19.600 & 0.200 &  -- \\
 & Std.Err     &  0.673 &  0.473 &   --   &  1.200 &   5.841 & 0.476 &  -- \\
 & T-value     &  0.594 & -2.328 &   --   & -1.333 &  -3.355 & 0.420 &  -- \\
\midrule
\multirow{3}{*}{Speech}
 & Coefficient & -1.300 & -1.700 & \textbf{0.300} & -11.000 & 0.800 &  --  &  -- \\
 & Std.Err     &  0.191 &  0.936 & \textbf{0.100} &   7.127 & 1.848 &  --  &  -- \\
 & T-value     & -6.789 & -1.816 & \textbf{3.000}$^{*}$ & -1.543 & 0.433 & -- & -- \\
\midrule
\multirow{3}{*}{Vision}
 & Coefficient & \textbf{167.000} & 2.200 & \textbf{166.900} & \textbf{51.600} & \textbf{7.300} & 0.100 & 0.400 \\
 & Std.Err     & \textbf{40.635}  & 1.376 & \textbf{58.421}  & \textbf{16.104} & \textbf{2.084} & 0.412 & 0.231 \\
 & T-value     & \textbf{4.110}$^{*}$ & 1.599 & \textbf{2.857}$^{*}$ & \textbf{3.204}$^{*}$ & \textbf{3.503}$^{*}$ & 0.243 & 1.732 \\
\bottomrule
\end{tabular}

\vspace{0.5em}
\raggedright
\footnotesize *Significance level of $p<0.05$. $\dagger$Significance level of $p<0.001$.
\end{minipage}
\end{table}
\renewcommand{\arraystretch}{1.0}

Figure \ref{fig5}(a)--(e) illustrate the trends of AI x topic technologies over the past five years (2014--2018). Figure \ref{fig5}(a) presents the trend for hardware-related topics. Figure \ref{fig5}(b) shows the trend for ML-related topics, with a notable increase in T2, which corresponds to deep learning technology. Figure \ref{fig5}(c) and Figure \ref{fig5}(d) display the topic trends for NLP and Speech, respectively. Finally, vision-related topics are shown in Figure \ref{fig5}(e), where Vision x \{T0, T2, T3\} technologies exhibit an upward trend over the past five years.

\texttt{DiTTO-LLM} applies statistical tests to trends in AI x topic technology to identify topic-based technology opportunities. Hypothesis testing determines whether the observed trend in AI x topic technology growth is temporary or statistically significant. For this analysis, the experiment employs equation (3), using the statistics from equation (2) to conduct the statistical hypothesis test.

Table \ref{table6} shows the results of a statistical test on the trend of AI x topic over the past 5 years (2014--2018) obtained from the matching dataset. In the table, the coefficient refers to equation (2). And Std.Err represents the standard error for the coefficient. Lastly, the T-value is the threshold for the probability that the coefficient is greater than 0.

The hypothesis test results indicate that, for ML, topics T0 and T2 reject the null hypothesis of equation (3) at a significance level of 0.05, suggesting that ML x \{T0, T2\} can be considered technology opportunities as of 2018. Similarly, Speech x T2 and Vision x \{T0, T2, T3, T4\} exhibit statistically significant trends, indicating their potential as technology opportunities. A detailed interpretation is provided in the next chapter.

\subsection{Module (b)+(c): TOD in 2019-2023} % 4.3
Module (b) of \texttt{DiTTO-LLM} extracts a new dataset from the PatentsView database, comprising 1,685,109 patents filed over the five-year period starting in 2019. The fine-tuned LLM from module (a) then inferred AI labels for these patents. Based on the inference results, we identified 529,649 samples with an AI probability of 0.5 or higher, which were defined as the new dataset for further analysis.

\renewcommand{\arraystretch}{1.2}
\begin{table}[H]
\centering
\begin{minipage}{0.8\textwidth}
\centering
\caption{List of topics extracted from new dataset}
\label{table7} % Table 9 -> Table 7
\sisetup{table-number-alignment = center, table-format=3.2} % align float
\begin{tabular}{clS}
\toprule
\textbf{Topic} & \textbf{Description} & \textbf{Rate (\%)} \\
\midrule
T0 & Serverless Architecture                   & 39.50 \\
T1 & Advanced Vision-to-Speech Technology      & 27.28 \\
T2 & Edge AI                                   & 13.81 \\
T3 & Information Retrieval and Semantic Search & 11.87  \\
T4 & Speech-to-Text Transcription Software     & 6.05  \\
T5 & SiC and GaN Power Semiconductors          & 1.48  \\
T6 & AI for Neuroprosthetics                   & 0.01  \\
\bottomrule
\end{tabular}
\end{minipage}
\end{table}
\renewcommand{\arraystretch}{1.0} % return to default

Table \ref{table7} presents the results of fine-tuned LLM-based topic modeling applied to the new dataset. Topic names were defined using the prompts in Table \ref{table2} and Table \ref{table3}. The topics extracted from the new dataset can be categorized into technologies for AI applications (T1, T3, T4), technologies for commercialization (T0, T2), and specialized technologies, including neuroscience (T5, T6).

Figure \ref{fig4}(b) displays the mapping results of topics and AI labels for the new dataset. As a result, Vision x T0 and Speech x T1 are likely saturated. In contrast, \{Hardware, Speech, Vision\} x T2 demonstrates balanced quantitative growth, indicating a suitable level compared to other technologies.

\begin{table}[H]
\centering
\begin{minipage}{0.95\textwidth}
\caption{Results of statistical tests for new dataset}
\label{table8} % Table 10 -> Table 8
\renewcommand{\arraystretch}{1.2}
\begin{tabular}{clrrrrrrr}
\toprule
\textbf{Technology} & \textbf{Statistics} & \textbf{T0} & \textbf{T1} & \textbf{T2} & \textbf{T3} & \textbf{T4} & \textbf{T5} & \textbf{T6} \\
\midrule
\multirow{3}{*}{Hardware}
 & Coefficient & -4.500 & -0.600 & \textbf{388.600} & 2.700 & 3.500 & -44.200 & -- \\
 & Std.Err     &  3.035 &  0.879 & \textbf{22.572}  & 1.159 & 2.527 &  6.420  & -- \\
 & T-value     & -1.483 & -0.682 & \textbf{17.216}$^{\dagger}$ & 2.330 & 1.385 & -6.885 & -- \\
\midrule
\multirow{3}{*}{ML}
 & Coefficient & -0.300 & -0.500 & 0.300 & -116.200 & -0.800 & 0.000 & -- \\
 & Std.Err     &  0.755 &  0.379 & 0.814 &   9.116  &  0.653 & 0.200 & -- \\
 & T-value     & -0.397 & -1.321 & 0.368 & -12.747  & -1.225 & 0.000 & -- \\
\midrule
\multirow{3}{*}{NLP}
 & Coefficient & -2.100 & -0.600 & -0.100 &  2.700 & -0.300 &   --  & -- \\
 & Std.Err     &  1.082 &  1.848 &  0.513 &  1.935 & 31.104 &   --  & -- \\
 & T-value     & -1.941 & -0.325 & -0.195 &  1.396 & -0.010 &   --  & -- \\
\midrule
\multirow{3}{*}{Speech}
 & Coefficient & -0.700 & -165.900 & \textbf{401.100} & 8.300 & 10.900 & 1.400 & -- \\
 & Std.Err     &  5.543 &  115.869 & \textbf{18.379}  & 11.458 & 7.295 & 1.149 & -- \\
 & T-value     & -0.126 & -1.432   & \textbf{21.824}$^{\dagger}$ & 0.724 & 1.494 & 1.219 & -- \\
\midrule
\multirow{3}{*}{Vision}
 & Coefficient & -485.500 & -1.700 & \textbf{12.400} & -26.400 & -10.100 & 2.500 & 0.200 \\
 & Std.Err     & 188.989  &  8.224 & \textbf{2.928}  & 28.144  &  5.707  & 1.620 & 0.503 \\
 & T-value     & -2.569   & -0.207 & \textbf{4.235}$^{*}$ & -0.938 & -1.770 & 1.544 & 0.397 \\
\bottomrule
\end{tabular}

\vspace{0.5em}
\raggedright
\footnotesize *Significance level of $p<0.05$. $\dagger$Significance level of $p<0.001$.
\end{minipage}
\end{table}
\renewcommand{\arraystretch}{1.0}

The topic associated with novel semiconductors (T5) shows a strong connection to hardware, while the neuroscience-related topic (T6) is primarily associated with AI technology in vision applications.

\begin{figure}[H]
\centering
\includegraphics[scale=0.85]{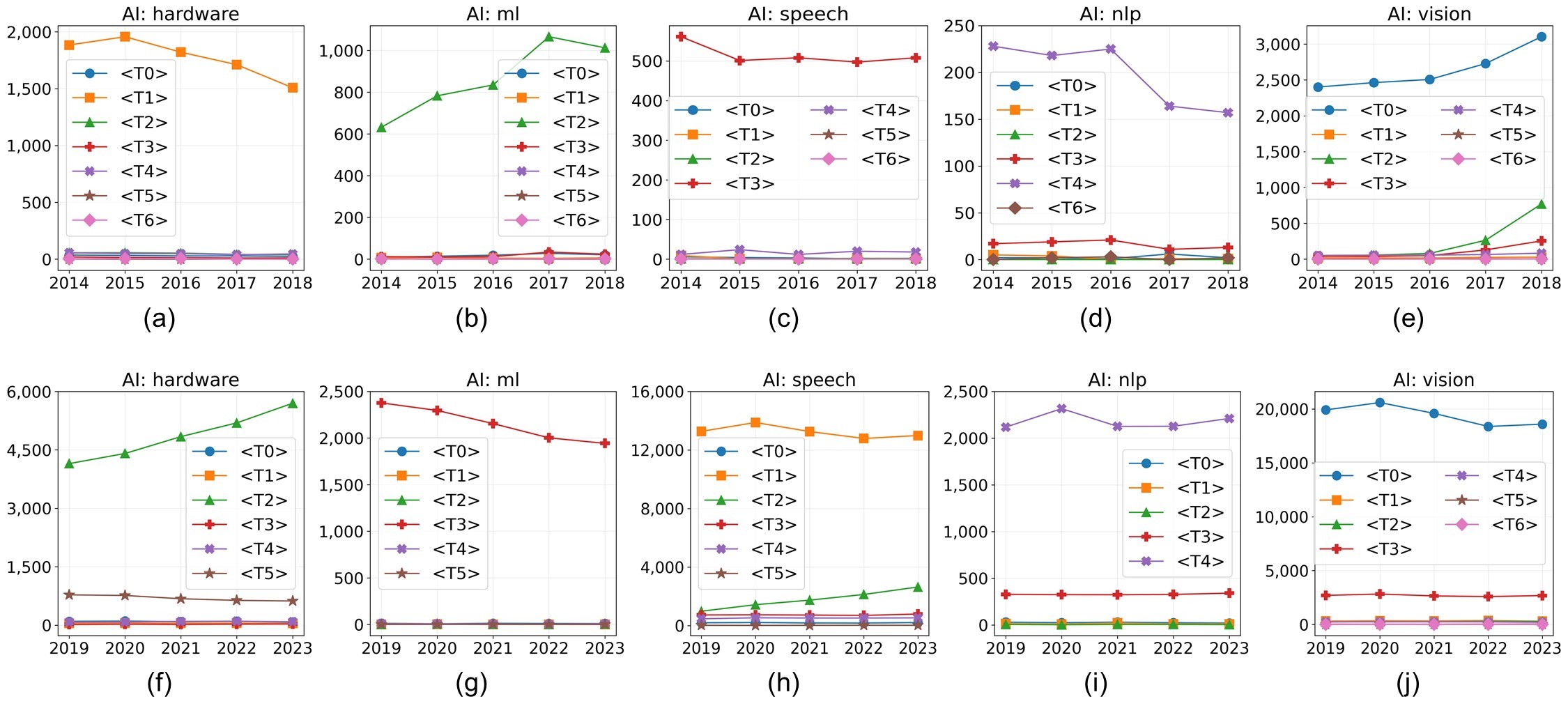}
\caption{Trends of AI x topic. (a)-(e): Results of matching dataset. (f)-(j): Results of new dataset.}
\label{fig5}
\end{figure}

Figure \ref{fig5}(f)--(j) illustrate the trends in AI x topic technologies over the past five years (2019--2023). Figure \ref{fig5}(f) and Figure \ref{fig5}(i) shows trends in topics related to Hardware and Speech, respectively. It can be seen that technologies related to Edge AI (T2) are growing rapidly in both hardware and speech. 

While other AI x topic technologies do not show notable growth, further analysis is needed to identify statistically significant technology opportunities through hypothesis testing.

Table \ref{table8} presents the results of statistical tests on the trends of AI x topic technologies over the past five years (2019--2023) from the new dataset. The hypothesis testing indicates that, between 2019 and 2023, Hardware, Speech, and Vision technologies related to Edge AI (T2) exhibit significant growth at the 0.05 significance level. These findings suggest that \{Hardware, Speech, Vision\} x T2 technologies represent future technology opportunities. A detailed interpretation is provided in the next chapter.

\section{Discussion} % Chapter 5
\label{ch5}
In experiment, we identified technology opportunities in the AI field using \texttt{DiTTO-LLM}. We now discuss the characteristics and contributions of these technology opportunities as revealed by the experimental results.

Technology opportunities are characterized by both quantitative growth and an upward trend over time. When identifying these opportunities, it is essential to focus on trends that show consistent increases over time. Technologies with merely quantitative growth may already be saturated, presenting challenges such as (i) limited potential for new business models and (ii) high barriers to market entry.

Returning to the experimental results, we identified AI x topics from the matching dataset (2000--2018) that are expected to represent technology opportunities. Empirical evidence for these opportunities is reflected in the mapping results of AI x topics. Specifically, by AI technology, Hardware x T1, ML x T2, NLP x T4, Speech x T3, and Vision x T0 are technologies that have shown quantitative growth (see Figure \ref{fig4}(a)).

However, Table \ref{table6} shows that the slope for Hardware x T1 was -99.8, indicating a decline over time. Similarly, both NLP x T4 and Speech x T3 are shown to be in decline. In contrast, ML x T2 (Deep Learning Technology), which has demonstrated quantitative growth, as well as ML x T0 (AI-based Image Recognition System), exhibit statistically significant upward trends.

Historically, advancements such as GoogleNet \cite{Szegedy2015}, ResNet \cite{He2016}, and SENet \cite{Hu2018} have significantly accelerated the development of image recognition technology using deep learning by 2018. Reflecting these advancements, Vision x T4 (Image-to-Text Converter) was identified as a technology opportunity. This indicates that the technology opportunities discovered by \texttt{DiTTO-LLM} are aligned with current drivers of the AI field, suggesting the validity and relevance of \texttt{DiTTO-LLM}'s findings in identifying appropriate technology opportunities.

Returning to the present, it is noteworthy that all technology opportunities identified in the new dataset (2019--2023) are associated with T2 (Edge AI). Previously, AI technologies primarily focused on algorithms for processing image and signal data. However, recent developments indicate rapid advancements in technologies aimed at commercializing these capabilities. In particular, \{Speech, Vision\} x T2 (Edge AI) holds the potential to significantly enhance user convenience by enabling efficient processing of image and signal data directly at the edge device.

Let us now compare past and present technology opportunities. Previously, there was a strong emphasis on AI technologies that could analyze diverse types of unstructured data. This demand included the need for technologies capable of processing not only speech and vision data but also multi-modal data that combines both modalities (see Vision x T3 in Table \ref{table6}). This trend likely reflected efforts to advance technologies such as the Image-to-Text Converter (see Vision x T4 in Table \ref{table6}).

In the future, numerous opportunities are anticipated to deliver AI-driven service functions to users. Edge devices—those closely connected to users—integrated with AI have the potential to greatly enhance daily convenience and comfort. Specifically, Edge AI combined with Speech and Vision (see \{Speech, Vision\} x T2 in Table \ref{table8}) shows promise for commercialization in areas such as smart energy cars, healthcare, and augmented reality \cite{Singh2023}. In this way, AI is progressing toward a more integrated presence in our everyday lives.

Are technologies focused on speech, vision, and edge AI indeed promising for the future? Can \{Speech, Vision\} x Edge AI open new industrial, scientific, and social opportunities? While we cannot predict the future with certainty, the empirical evidence obtained from \texttt{DiTTO-LLM} offers valuable insights. The technology opportunities discovered by \texttt{DiTTO-LLM} in 2018, such as ML and deep learning, still provide us with many opportunities. Based on this evidence, we can affirmatively answer the question, with the hope that these technologies will foster positive impacts across industry, science, and society.

\section{Conclusion} % Chapter 6
\label{ch6}
In today’s complex global environment, technology opportunities serve as a cornerstone for innovative growth for both countries and companies. Numerous studies have highlighted the importance of monitoring inter-technology relationships and tracking technological evolution over time to effectively identify emerging technology opportunities.

Technology opportunities manifest in various forms, including technological vacancies, convergent technologies, and emerging technologies. Previous studies have employed expert, time series, mapping, and text-based approaches for TOD. This paper proposes a TOD framework that leverages LLM-based topic modeling to address limitations of existing approaches.

The proposed framework extracts topics from the USPTO AIPD and maps them to AI technologies. This framework identifies technologies that exhibit quantitative growth within the AI x topic map and detects technologies with increasing trends over time.

This paper conducted two experiments using the USPTO AIPD. The first experiment aimed to evaluate the proposed framework by identifying past technology opportunities, while the second experiment focused on discovering future technology opportunities.

In the first experiment, we detected opportunities for AI technologies in vision through advancements in ML and deep learning. The second experiment suggested that, in the future, AI technology will likely evolve into more accessible forms, such as integration with edge devices, making it more readily available in everyday settings.

This paper proposed a framework for TOD through an LLM-based topic modeling approach. The framework successfully identified statistically significant TODs.

However, the limitations of this study are as follows:
\begin{itemize}
\item First, the proposed framework relies on the USPTO AIPD, which may lead to greater errors when forecasting technology opportunities for AI in the more distant future.
\item Second, the framework does not incorporate prompt engineering for topic definition, which can affect TOD interpretations, as results may vary based on how topics are defined.
\item Lastly, the framework is limited to patents and does not consider scientific documents, such as academic publications or conference proceedings, which are also valuable sources for discovering AI technology opportunities
\end{itemize}

In future research, incorporating prompt engineering to define topics more precisely is essential, as this is expected to yield more specific technology opportunity discoveries. Additionally, developing a TOD framework that integrates both patents and academic publications is necessary. Such an approach would enable the discovery of technology opportunities that leverage insights from both academia and industry.

\section*{Appendix} % Appendix
Table A1 presents the query $Q$ used to extract the AIPD from the PatentsView database for this experiment.

\renewcommand{\arraystretch}{1.4}
\begin{table}[H]
\centering
\begin{minipage}{0.9\textwidth}
\centering
\caption*{Table A1. Query to extract titles and abstracts from the PatentView DB}
\label{tableA1}
\begin{tabular}{@{}p{0.03\linewidth} >{\raggedright\arraybackslash}p{0.7\linewidth}@{}}
\toprule
\textbf{Step} & \textbf{Query} \\
\midrule
1: & \textbf{SELECT} \texttt{A.application\_id}, \texttt{A.patent\_id}, \texttt{A.filing\_date}, \texttt{P.patent\_date}, \texttt{P.patent\_title}, \texttt{P.patent\_abstract} \\
2: & \textbf{FROM} \texttt{g\_application A} \\
3: & \textbf{JOIN} \texttt{g\_patent P} \textbf{ON} \texttt{A.patent\_id = P.patent\_id} \\
4: & \textbf{WHERE} \texttt{A.application\_id} \textbf{IN}($d_{ap}$) \\
\bottomrule
\end{tabular}
\end{minipage}
\end{table}
\renewcommand{\arraystretch}{1.0} % return to default

The query used in the experiment utilizes two database tables. The first table, \texttt{g\_application}, contains basic information about the patent application. The second table, \texttt{g\_patent}, includes text fields such as the patent title and abstract.

%Bibliography
\bibliographystyle{unsrt}  
\bibliography{references}  

\end{document}